\documentclass[lettersize,journal]{IEEEtran}
\usepackage{amsmath,amsfonts}
\usepackage[utf8]{inputenc} 
\usepackage[T1]{fontenc}    
\usepackage{hyperref}       
\usepackage{url}            
\usepackage{booktabs}       
\usepackage{amsfonts, amsmath}       
\usepackage{nicefrac}       
\usepackage{microtype}      
\usepackage{xcolor}         
\usepackage{graphicx}
\usepackage{algorithmic}
\usepackage{algorithm}
\usepackage{array}
\usepackage{subcaption}
\usepackage{textcomp}
\usepackage{stfloats}
\usepackage{url}
\usepackage{verbatim}
\usepackage{graphicx}
\usepackage{cite}
\usepackage{bm} 
\usepackage{newfloat}
\usepackage{listings}
\usepackage{makecell}
\usepackage{enumitem,kantlipsum}
\usepackage{empheq}
\usepackage{wrapfig}
\usepackage{natbib}
\usepackage{scalerel}
\usepackage{multirow}
\usepackage{booktabs}

\begin{document}

\title{
  Prioritized League Reinforcement Learning for Large-Scale Heterogeneous Multiagent Systems

}
\author{IEEE Publication Technology,~\IEEEmembership{Staff,~IEEE,}
\thanks{This paper was produced by the IEEE Publication Technology Group. They are in Piscataway, NJ.}
\thanks{Manuscript received April 19, 2021; revised August 16, 2021.}}

\markboth{Journal of \LaTeX\ Class Files,~Vol.~14, No.~8, August~2021}%
{Shell \MakeLowercase{\textit{et al.}}: A Sample Article Using IEEEtran.cls for IEEE Journals}


\maketitle

\begin{abstract}
   Large-scale heterogeneous multi-agent systems consider various realistic factors present in the real world, such as agents with diverse abilities and overall system cost. In comparison to homogeneous systems, heterogeneous systems offer significant practical advantages. Nonetheless, they also present challenges for multi-agent reinforcement learning, including addressing the non-stationary problem and managing an imbalanced number of agents with different types.
We propose a Prioritized Heterogeneous League Reinforcement Learning (PHLRL) method to address large-scale heterogeneous cooperation problems.
PHLRL maintains a record of various policies that agents have explored during their training, establishing a league consisting of diverse policies to aid in future policy optimization. 
Furthermore, we introduce prioritized advantage coefficients to compensate for the gap caused by differences in the number of different types of intelligent agents.
Next, we use Unreal Engine to design a Large-Scale heterogeneous Cooperation benchmark name LSHC, a complex attack scenario that requires collabration from both ground and airbone agents.
We use empirical experiments to show that PHLRL outperforms SOTA methods including Qmix, Qplex and cw-Qmix in LSHC.

\end{abstract}


\section{Introduction}

Large-scale heterogeneous multiagent system is commonly seen in the natural world.
For example,
each ant individual in an colony has very different biological forms and social roles depending on its division of labor.
This heterogeneous characteristic enables agents to efficiently optimize their collective success with minimal resource requirements.
Comparing with homogeneous systems, 
these systems is featured by a combination of multiple different types of agents. 
Different agents in homogeneous systems serve very different functions, 
their capability and structures vary to accommodate these differences. 
Furthermore, the heterogeneous systems can be extended greatly in application because it has lower cost and 
greater flexibility compared to homogeneous systems.
Additionally, by leveraging the individual strengths of various agents, tasks can be distributed more evenly, 
reducing the burden on any single agent and increasing the system's overall efficiency.

Many recent classic Multiagent Reinforcement Learning (MARL) algorithms 
do not consider the ability distinction between agents explicitly. 
They primarily focuses on either homogeneous environments and have been proved effective in solving these tasks. 
However, in a diverse world, relying solely on homogeneous systems is not sufficient to efficiently solve complex teamwork problems. 
Real multiagent systems often require diverse agents to participate in order to ensure division of labor and reduce costs. 
When applying MARL algorithms to heterogeneous multiagent tasks, there are two main challenges that need to be addressed:

(1) Heterogeneous non-stationarity problem.
Non-stationarity \cite{hernandez2017survey, fu2022learning} significantly influences the performance of heterogeneous multiagent systems, especially large-scale systems.
Unlike homogeneous multiagent systems with identical agents,
a large-scale heterogeneous system contain multiple types of agents that vary in ability, quantity, value, and action space.
The inequality of the number of different agent types creates gap for cross-type agent cooperation in the process of reinforcement learning.
Furthermore, essential agent types are usually fewer in number in heterogeneous systems, 
e.g., queens in a hive of bees and AWACS aircrafts in military operations.
The importance of these agents does not align with their number in multiagent systems. 
In extreme cases, just one of these essential agents is sufficient for completing a designated task.
However, 
these essential agents suffer from the high variance in state-value estimation due to insufficient sampling,
which can destabilize the entire team learning process.

(2) Decentralized large-scale deployment problem. Heterogeneous systems have many realisitic concerns compared with homogeneous systems.
Firstly, agent-wise communication in large-scale heterogeneous systems is limited by brandwidth and information processing capability of individual agents,
as a result, the decentralized execution paradigm is prefered in large-scale systems with hundreds of agents.
Secondly, the robustness of agent policies is highly valued in heterogeneous environments 
because agents can suffer from degenerated performance when external environment or agent properties deviates from training baselines.


To address the challenges associated with heterogeneous large-scale multiagent systems, 
we propose a novel Prioritized Heterogeneous League Reinforcement Learning (PHLRL) method that can effectively solve general cooperation problems under diverse and varying conditions. 
In addition, we have developed a Large-Scale Heterogeneous Cooperation (LSHC) benchmark using the Unreal Engine platform to efficiently simulate large-scale multiagent cooperation scenarios. 
Our extensive experiments show that the PHLRL method exhibits superior performance and effectively addresses complex heterogeneous multiagent cooperation in LSHC.
We use empirical experiments to show that PHLRL outperforms SOTA methods including Qmix, Qplex and cw-Qmix in LSHC.

In this paper, our contributions are as follows:
1. We propose Heterogeneous League Training (xxx), 
a general-purpose MARL algorithm for heterogeneous multiagent tasks.
2. We demonstrate the superior performance of xxx and the effectiveness of xxx in addressing the policy version iteration problem.
3. Based on xxx, we propose a method for evaluating
the difficulty of learning the roles undertaken by different agent types.
\section{Related Works}

Multiagent system has attracted great interests in recent years in the field of reinforcement learning,
it has several different perspectives considering the task scenario.
A great proportion of studies do not consider the characteristic of heterogeneous multiagent systems. 
VDN \cite{sunehag2017value} algorithm improves the original Independently Q-Learning (IQL) \cite{watkins1989learning} model and addresses general multiagent cooperation problems with reward decomposition.
Qmix \cite{rashid2018qmix} and its successor models \cite{rashid2020weighted} proposes an alternative mix approach that satisfy the Individual Global Max (IGM) principle but can represent a much richer class of Q functions.

Some studies with explicit heterogeneous modeling contribute to problems in specific domains. \citep{yu2019deep} investigate an RL-based protocol for heterogeneous wireless networking. \citep{ishiwaka2003approach} address pursuit problems where hunter agents have different abilities.
\citep{zhao2019deep} use heterogeneous system to model cellular mobile networks and propose a promising technique to reduce reduce the deployment costs.
\citep{orhean2018new} put forward an algorithm to solve the scheduling problem in distributed cloud computing systems.
Furthermore, the heterogeneous multiagent RL algorithms are also developed in vehicle routing \cite{li2021deep, qin2021novel}, delivery system \cite{li2021heterogeneous}, satellite communication \cite{deng2019next, jiang2020reinforcement}, etc.
Moreover, there are numerious studies that investigate heterogeneous multiagent systems without reinforcement learning.
\citep{bao2022recent} conduct a survey which discusses many recent achievement on heterogeneous multi-agent systems with various constraints, such as parameter uncertainties, disturbances, etc.
Classic methods such as leader-follower \cite{gong2021bounded, yan2021flocking}, sliding mode control \cite{liu2019decentralized, ye2018consensus}, etc.

General-purpose multiagent RL algorithms for heterogeneous systems are relatively fewer in number. 
An early study \cite{kapetanakis2004reinforcement} explored a 2-player heterogeneous table games with Q-learning. \citep{wakilpoor2020heterogeneous} improve situational awareness by utilizing teams of unmanned aerial vehicles.
\citep{fu2022learning} address the policy iteration problem in heterogeneous RL problems.
However, these general-purpose algorithms are limited to small-scale multiagent settings and has not consider the challenges in large-scale scenarios.

Most heterogeneous RL algorithms are tested under various simulation environments. Some RL simulation environments used for multiagent studies primarily focus on tasks with homogeneous agents \cite{deka2021natural, zheng2018magent, fu2022concentration}. However, there is an emerging trend of benchmark environments that involve cooperation among heterogeneous agents. One such example is the Multi-Agent Particle Environment (MAPE) \cite{lowe2017multi}, which is designed for less complex tasks like predator-prey and speaker-listener scenarios. Another benchmark environment, SMAC \cite{samvelyan2019starcraft}, showcases heterogeneous team combinations such as 1c3s5z and MMM2. Nonetheless, only a small portion of the maps (e.g., MMM2) consider the supportive relationships among team agents. And few benchmark environment can simulate realistic large-scale systems participated by large-scale heterogeneous agents.
\section{Preliminaries}

\subsection{Heterogeneous Multiagent System.}
A heterogeneous multiagent system $H$ can be modeled by a Dec-POMDP \cite{oliehoek2016concise} formulated by 
$H = \langle A, \varDelta, \mathbf{U}, \mathbf{S}, P_\mathbf{s}, \mathbf{O}, P_\mathbf{o}, R, \gamma\rangle$.

$A=\lbrace a_{1}, \dots, a_{N} \rbrace$ is the set of heterogeneous agents and $N=|A|$ is the total number of agents.
Each agent differs with each other,
$\varDelta = \lbrace d_1, \dots, d_{M} \rbrace$ is the collection of all agent types in the system, $M=|\varDelta|$ is the total number of agent types.
Corrispondingly, the type identity of each agent $a_i$ if denoted as $\operatorname{TP}(a_i) \in \varDelta$.
It is important to note that the number of agent types $M$ is usually significantly smaller than total agent number $N$ in most large-scale heterogeneous systems, $M \ll N$. 
$\mathbf{U}=\mathcal{U}_1 \times \dots \times \mathcal{U}_N$ is the joint action space, where $\mathcal{U}_i$ represents the action space for each agent $a_i$ and $\mathbf{u}_t \in \mathbf{U}$ is joint action sampled from $\mathbf{U}$ at time step $t$.
$\mathbf{S}$ is the global state space. Environment state can be denoted as $\mathbf{s}_t \in \mathbf{S}$, and $P_\mathbf{s}(\mathbf{s}_{t+1}|\mathbf{s}_t, \mathbf{u}_t)$ is a stochastic state transition function.
The observation space of agents is denoted as $\mathbf{O}=\mathcal{O}_1 \times \dots \times \mathcal{O}_N$ and the joint observation of agents is denoted as $\mathbf{o}_t \in \mathbf{O}$. Similarly, $\mathcal{O}_i$ represents the observation space for each agent individual $a_i$.
Due to the partially observable constraint and possible random disturbances, 
the observation of agents is determined by another stochastic function
$P_\mathbf{o}(\mathbf{o}_t|\mathbf{s}_t)$.
Finally, $R(\mathbf{s}_t)$ is the reward function and $\gamma$ is the discount factor.

\subsection{League Training.}

In AlphaStar, a novel approach to reinforcement learning, agents are trained through a combination of supervised learning from expert data and reinforcement learning. This results in a diverse population of policies. 
During Reinforcement Learning training, agents intermittently create duplicate versions of themselves and store these copies as past players.
their past versions frozen. They are trained against each other as well as their own past selves. This approach enables agents to adapt to the constantly evolving strategies of opponents and improves model performance in heterogeneous Multi-Agent Reinforcement Learning (MARL) scenarios.

However, league training is a resource-consuming method that requires a large cluster of computing nodes.
To address this drawback,
\citep{ma2021distributed} propose a smaller scale league training method for RTS game without human knowledge,
and \citep{fu2022learning} apply league training method to small-scale heterogeneous reinforcement learning.



\section{Prioritized Heterogeneous League Reinforcement Learning}

We propose a Prioritized Heterogeneous League Reinforcement Learning (PHLRL) method to address the large-scale heterogeneous cooperation problems.

\begin{figure*}[!t]
  \centering
  \includegraphics[width=\linewidth]{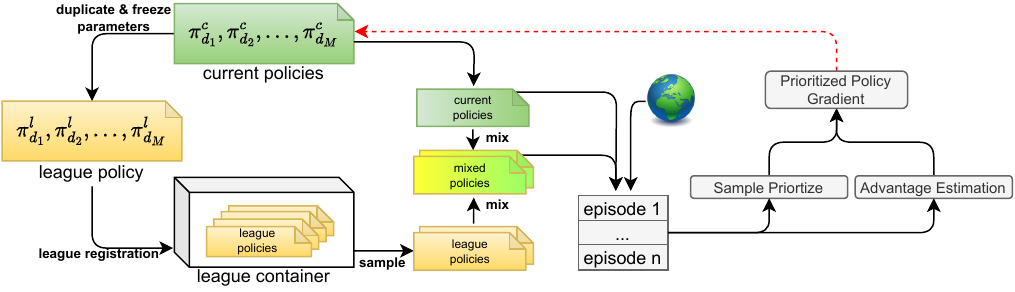}
  \caption{A high-level illustration of the training procedure of PHLRL method. In each iteration, the frontier agent policies are mixed with historical league policies. After agents explore the large-scale heterogeneous environment, the sample of each episodes are prioritized according to the relative performance of each agent type. Finally, the frontier agent policies are optimized with the prioritized policy gradient calculated from these samples.
}
\label{fig:main}
\end{figure*}

On the whole, PHLRL addresses the large-scale heterogeneous multiagent reinforcement learning problem by:
\begin{enumerate}
  \item Learning robust cooperation policies by cooperating with teammates with multifarious policies.
  \item Resolving the sample inequality problem between different agent types.
\end{enumerate}

\subsection{Heterogeneous League.}

Building a league to contain a collection of policies is an effective way for producing multifarious policies, as it is investigated by \cite{vinyals2019grandmaster, fu2022learning}.

The early applications of league training are proposed by \cite{vinyals2019grandmaster}.
The adoption of the league training technique allows agents to enhance their policies by learning from both the experiences of working together with a stable team and collaborating with external agents who have different sets of cooperation skills. 
This technique effectively tackles the issue of learning instability by encouraging agents to adapt their cooperation strategies according to the individual characteristics of their teammates.

In this study, we assume that same-type agents share the same policy.
We define a policy group (denoted as $\mathbf{\Pi}_\phi$) to represent all policies of all agent types in a heterogeneous team:
$$
  \mathbf{\Pi}_\phi = \lbrace \pi_{d_1},\dots, \pi_{d_M} \rbrace,
$$
where $\lbrace \pi_{d_1},\dots, \pi_{d_M} \rbrace$ is the policies of agents with type $\lbrace d_1, \dots, d_M \rbrace$ respectively, and the parameters of the policy group is represented by $\phi = \lbrace \phi_{d_1}, \dots, \phi_{d_M} \rbrace$.

During the heterogeneous training process, the most important policy group is the frontier policy group $\mathbf{\Pi}^f$. The frontier policy group $\mathbf{\Pi}^f$ is the only policy group that is continuously trained and updated. To this end, our method distincts from the AlphaStar method (AlphaStar actively trains all policies unselectively). Limiting the number of training policies can reduce the computational cost and accerate the learning process.

Next, the league $\mathcal{L}$ in this work is defined as a container that stores past and policy groups.
$$
\mathcal{L} = \lbrace \mathbf{\Pi}^1_{\phi_{1}}, \dots, \mathbf{\Pi}^K_{\phi_{K}} \rbrace,
$$
where the size of the league is denoted as $K$.
League $\mathcal{L}$ has a capacity limit $K_m$ and $|\mathcal{L}|<=K_m$.
All league policy groups are duplication of the frontier policy group $\mathbf{\Pi}^f$ frozen at a specific iteration step, and the league policies are frozen once they are stored in $\mathcal{L}$.
On this particular matter, 
our method differs from \cite{vinyals2019grandmaster}, 
which does not freeze the parameters of league policies to further maximize the competitiveness.
However, the approach proposed by \cite{vinyals2019grandmaster} is very computationally expensive, 
and is hard to scale to large-scale multiagent cooperation tasks.
As this work concentrates only on the cooperation problem in large-scale hetegeneous cooperation,
the league is constructed very differently.

Next, we introduce PHLRL method, beginning from random initial policies and empty league.

\subsection{Learning Robust Heterogeneous Policies.}

The PHLRL method begins by initializing random frontier policies $\mathbf{\pi}_{\text{f}}$ and an empty league $\mathcal{L} = \lbrace \rbrace$.
Each policy iteration has following stages:
\begin{enumerate}
  \item Generate mixed policy combinations from frontier policy group and league policy groups. (If the league is not empty $|\mathcal{L}| != 0$)
  \item Perform agents-environment simulation. Gather samples from $N_E$ simulated episodes.
  \item Calculate sample priority according to the relative performance of each agent type.
  \item Compute the prioritized policy gradient.
  \item Update the frontier policy group with the prioritized policy gradient.
  \item Duplicate and add the new frontier policy group to the league if it passes the evaluation of the league.
\end{enumerate}

Beginning from the first stage, before each episode a mixed policy combination if sampled from the frontier policy group and the league policy groups.
A mixed policy is a hybridization of frontier and league policies when the league is not empty.
However, the league is empty at early training iterations.
In such case, only the frontier policy group is used for agent-environment simulation.
To generate a mixed policy combination,
a policy group $\mathbf{\Pi}^k$ is sampled from the league with a uniform distribution:
$$
  \mathbf{\Pi}^k \sim U(\mathcal{L}), 1 \le k \le K
$$
Next, one of the agent types $d_x$ is sampled from the policy group:
$$
  d_m  \sim U(\varDelta)
$$
Then, the agents of type $d_m$ are assigned with policy $\pi^{l_k}_{d_m} \in \mathbf{\Pi}^k$. Let $d_{m^-}$ be any agent types that are not $d_m$, other agents will execute the corrisponding frontier policies $\mathbf{\Pi}^{f}_{d_{m-}} \subset \mathbf{\Pi}^{f}$,
where $\mathbf{\Pi}^{f}_{d_{m-}} = \mathbf{\Pi}^f - \lbrace \pi^{f}_{d_m} \rbrace$.
This mixed policy is effective for only one episode.
When a new episode begins, a new mixed temporary policy combination $\mathbf{\Pi}^{[k,d_m]}$ will generated, where $\mathbf{\Pi}^{[k,d_m]}= \mathbf{\Pi}^{f}_{d_{m-}} \cup \lbrace \pi^{k}_{d_m} \rbrace$.
Generating different policy combinations within a policy iteration
makes it possible to observe and analyze the relative performance differences of each agent type.
Furthermore, the differences of performance level are reliable indicators for calculating the sample priority.

The same policy sampling process is repeated $N_E$ times, and each generated mixed policy group is used to perform agents-environment simulation that last for a single episode. Therefore, $N_E$ episodes are simulated in a iteration, the sample of which is denoted as $\mathcal{D}$.

\subsection{Prioritized Policy Gradient.}

The learning objective with league $\mathcal{L}$ is to maximize the expected return:
$$
J^{\mathcal{L}} = \mathbb{E}_{\tau_t \sim \mathcal{D}}[R(\mathbf{s}_t)],
$$
where $\tau_t$ is a piece of sample from the replay buffer $\mathcal{D}$, and $\tau_t(i, k, d_m) = \lbrace s^i_t, o^i_t, u^i_t, r^t, k, d_m \rbrace$.

In PHLRL the performance of a policy group $\mathbf{\Pi}$ is defined by its win rate, denoted as $\beta(\mathbf{\Pi})$.

And $\beta^{[d_m]}$ is represented as the averaged performance of policies of $d_m$ agents combined with frontier policies.
$$
\beta^{[d_m]} \doteq \mathbb{E}_{ \mathbf{\Pi}^k \sim U(\mathcal{L})} \beta(\mathbf{\Pi}^{[k,d_m]})
$$
The overall performance ${\beta^{[\Delta]}} $ is the performance of league and frontier policies at current iteration:
$$
{\beta^{[\Delta]}} \doteq \mathbb{E}_{ d_m  \sim U(\varDelta) } \beta^{[d_m]}
$$
Finally, for a trajectory $\tau_t(i, k, d_m) \in \mathcal{D}$, the advantage prioritization factor is defined as:
$$
A_{\beta}\left[\tau_t(i, k, d_m)\right] \doteq \frac{\psi + {\beta^{[\Delta]}}}{\psi + \beta^{[d_m]}},
$$
where $\psi \in (0,1)$ is a constant.

The advantage estimation can be formulated by GAE \cite{schulman2015high}:
$$
A_{{GAE}}(\tau_t) = \sum^{\infty}_{l=0} (\gamma \lambda)^t \delta_{t+l}({\tau_t})
$$

$$
\delta_{t+l}({\tau_t}) = r_t + \gamma \cdot V(s^i_{t+1}) - V(s^i_t)
$$

The final Prioritized Policy Gradient is formulated by:
$$
\nabla_{\phi^f}(J^{\mathcal{L}}) 
= \mathbb{E}_{\tau_t(i, k, d_m) \in \mathcal{D}}
\left[
  \nabla_{\phi^f} \log
    \Pi^{[k, d_m]}_{\phi^\mathcal{L},\phi^f}(u^i_t|o^i_t)
    \cdot
    A_{{GAE}}(\tau_t) A_{\beta}(\tau_t)
  \right]
$$ 

This study uses PPO to optimize the policy nerual network, consequently,
the gradient formulation has to be re-formulated by loss $L^{CLIP}(\phi)$:
$$
w_t(\phi^i) = \frac{\pi_{\phi^i}(u_t|o_t)}{\pi_{\phi^i_{\text{old}}}(u_t|o_t)}
$$
$$
w^{CLIP}_t(\phi^i) = \min\left(w_t(\phi^i), \text{clip}\left(w_t(\phi^i), 1 - \epsilon, 1 + \epsilon\right)\right)
$$
$$
L^{CLIP}(\phi) = \hat{\mathbb{E}}_{
        \tau_t(i, k, d_m) \in \mathcal{D}
    }  \left[ w^{CLIP}_t(\phi^i) A_{{GAE}}(\tau_t) A_{\beta}(\tau_t)\right]
$$

As mentioned previously, the frontier policy group is the only policy group that is continuously trained, while league policy parameters are frozen. Thereby, only frontier policies parameters $\lbrace \phi^1, \dots, \phi^n \rbrace$ are updated with the prioritized policy gradient.

\subsection{League Update}
The goal of league update is to maintain a league that contains a collection of policies that are competitive and diverse. 
The league in PHLRL starts with an empty league $\mathcal{L} = \lbrace \rbrace$. 
New league member is forked and duplicated from the frontier policy group once every $N_{\text{u}}$ PPO iterations.
The process is specifically divided into the following steps:
\begin{enumerate}
  \item Duplicate the frontier policy group $\mathbf{\Pi}^f=\lbrace \pi^f_1, \dots, \pi^f_{d_M} \rbrace$, denote the duplication as $\mathbf{\Pi}^{K+1}$.
  \item Add $\mathbf{\Pi}^{\mathcal{L}\text{new}}$ to the league $\mathcal{L}$.
  \item Sort the league according policy group performance $\lbrace \beta(\mathbf{\Pi}^1) \dots, \beta(\mathbf{\Pi}^K), \beta(\mathbf{\Pi}^{K+1}) \rbrace$.
  \item If the league exceeds its capacity $K_m$, i.e. $K+1 > K_m$, locate the two groups $\beta(\mathbf{\Pi}^p)$, $\beta(\mathbf{\Pi}^q)$ with the minimal performance differences: 
  $$
  |\beta(\mathbf{\Pi}^p)-\beta(\mathbf{\Pi}^q)| \le |\beta(\mathbf{\Pi}^{p'})-\beta(\mathbf{\Pi}^{q'})|,  \forall \mathbf{\Pi}^{p'} \in \mathcal{L} , \mathbf{\Pi}^{q'} \in \mathcal{L},
  $$
  then remove either $\beta(\mathbf{\Pi}^p)$ or $\beta(\mathbf{\Pi}^q)$.
  Considering training stability, it is further specify that the newer policy group is removed. For instance, when the latest $\mathbf{\Pi}^{K+1}$ policy group happens to be $\beta(\mathbf{\Pi}^p)$ or $\beta(\mathbf{\Pi}^q)$, it is removed instead of the other.
  \item Proceed another $N_{\text{u}}$ PPO iterations and repeat from 1).
\end{enumerate}

\subsection{Adaptive Hypernetwork-Based Neural Policy}

Collaborating with various partners having different policies can be a challenging job. While some partners may be highly skilled and cooperative, others may be weak, with immature policies drawn from the league. Consequently, it becomes crucial to provide agents with a neural network setup that is capable of identifying and addressing these disparities in partners. 
This section investigated about how to design policy neural networks that can:
\begin{enumerate}
\item adapt to different agent types.
\item adapt to different policy combinations.
\item adapt to the transition between frontier policies and league policies.
\end{enumerate}

\begin{figure*}[!t]
  \centering
  \includegraphics[width=0.6\linewidth]{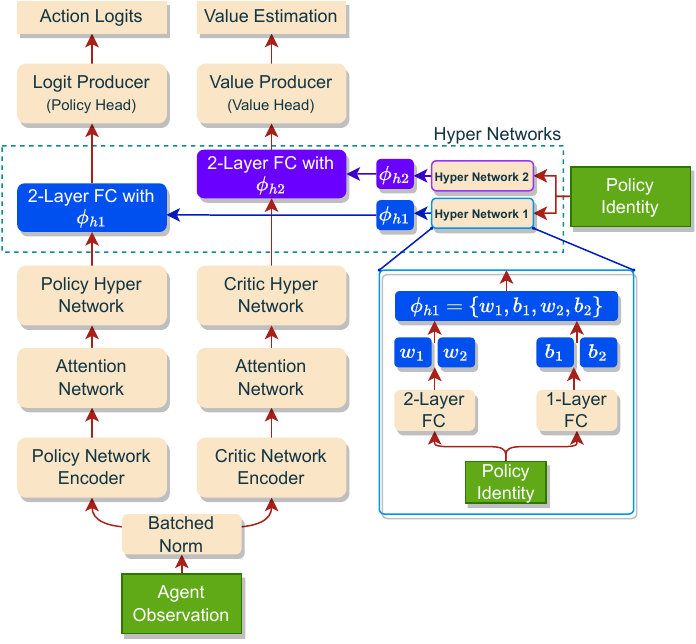}
  \caption{
    The policy and critic neural network structure of PHLRL method. The key difference with regular agent policies is the application of hyper networks. 
  }
\label{fig:hyper}
\end{figure*}

To address these three problems simultaneously, we propose a hypernetwork-based neural policy architecture.
As illustrated in Fig.~\ref{fig:hyper}, the policy network and the critic network is established with two types of networks: the regular feed-forward networks and the hypernetworks (inside the dotted rectangle).

Both the policy network and the critic network requires the agent observation representation and a policy identity representation to determine next move or estimate current state value.

\begin{enumerate}
\item Agent observation representation $\mathbf{o}_t$: the observation of each agent.

\item Policy identity representation $F_t$: the awareness of the difference of policy group combinations is the prerequisite for achieving adaptive collaborative behaviors. In other words, agents need to acquire the information of their current policy group combination. Policy identity representation $F_t$ is a vector that encodes the information of the current policy group combination.
\end{enumerate}

The composition of policy identity representation $F_i(d_i, \mathbf{\Pi}^{[k, d_m]})$ for agent $i$ is defined as follows:
\begin{equation}
\begin{aligned}
  F_i(d_i, \mathbf{\Pi}^{[k, d_m]}) = &\operatorname{concat} \lbrace  \\
  & \quad \left[ f(d_x | d_i, [k, d_m]) \mid d_x \in \left\lbrace d_1, \dots, d_M \right\rbrace \right], \\
  & \quad \operatorname{onehot}(d_i)  \\
  & \rbrace \\
\end{aligned}
\end{equation}

\begin{equation}
  f(d_x | d_i, [k, d_m])=\left\{\begin{array}{ll}
    \beta(\mathbf{\Pi}^k), & \text{if }  d_x = d_m \text{ and } d_i \neq d_m \\
    \\
    \beta^f, & \text{otherwise}
\end{array}\right. ,
\end{equation}
where $\mathbf{\Pi}^{[k, d_m]}$ is the current policy combination that is composed by league policy $\pi^{k}_{d_m}$ and frontier policies $\mathbf{\Pi}^{f}_{d_{m-}}$. $\mathbf{\Pi}^{[k,d_m]}= \mathbf{\Pi}^{f}_{d_{m-}} \cup \lbrace \pi^{k}_{d_m} \rbrace$. And $d_i$ is the agent type of agent $i$. And $\beta^f=1$ is a placeholder constant.

For instance, the representation of $F_i(d_i, \mathbf{\Pi}^{[k, d_m]})$ can describe the team combination. 
For example, for a heterogenous team with 4 types of agents $\Delta=\lbrace d_1, d_2, d_3, d_4 \rbrace$, three agents $a_1, a_2, a_3$ belong to type $d_1, d_2, d_3$ respectively. In one of the training episode, the current policy combination is sampled as $\mathbf{\Pi}^{[k, d_2]}$. In this case, policy identity representation of $a_1$ is $F_1(d_1, \mathbf{\Pi}^{[k, d_2]}) = \left[ \beta^f, \beta(\mathbf{\Pi}^k), \beta^f, \beta^f,\ \  1, 0, 0, 0 \right]$, that of $a_2$ is $F_2(d_2, \mathbf{\Pi}^{[k, d_2]}) = \left[ \beta^f, \beta^f, \beta^f, \beta^f,\ \  0, 1, 0, 0 \right]$, then that of $a_3$ is $F_3(d_3, \mathbf{\Pi}^{[k, d_2]}) = \left[ \beta^f, \beta(\mathbf{\Pi}^k), \beta^f, \beta^f,\ \  0, 0, 1, 0 \right]$.

\section{Experiments}
\begin{figure*}[!t]
  \centering
  \includegraphics[width=0.8\linewidth]{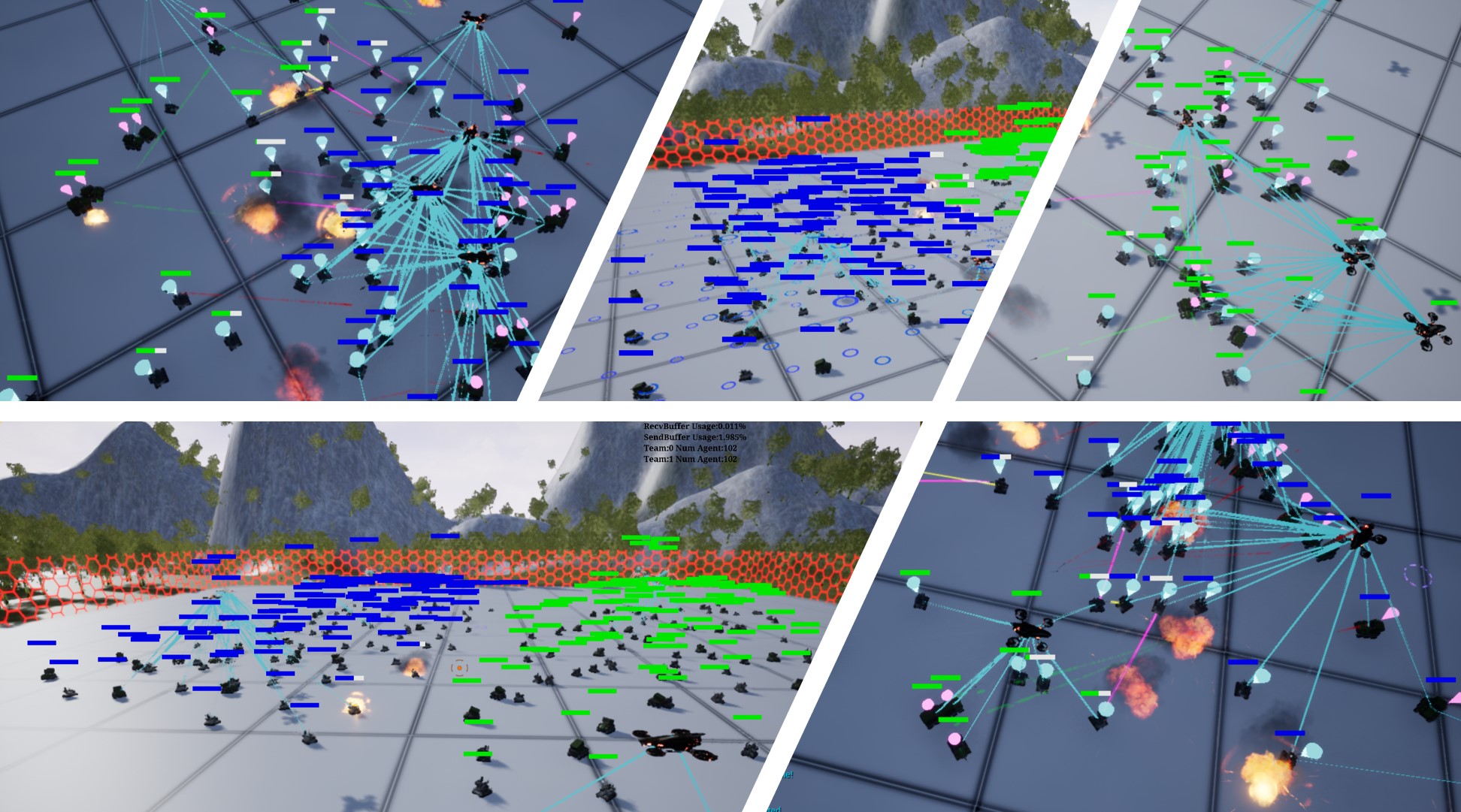}
  \caption{
    The Large Scale multiagent OPeration (LSOP) benchmark environment.
  }
\label{fig:LSOP}
\end{figure*}

\subsection{Heterogeneous Benchmark Environment.}

To investigate the performance of PHLRL, we adopt Unreal-MAP \cite{fu2023unrealmap}, a project for creating complex heterogeneous multiagent task environments, to build a simulation environment. As shown in Fig.~\ref{fig:LSOP}, the task environment we have created a large-scale simulation task environment that simulates air and ground operations, named as Large Scale multiagent OPeration (LSOP). The simulation parameters are listed in Table.~\ref{tab:env}.

There are three types of agents in LSOP: missile vehicles, gun vehicles and air drones. 
Missile vehicles are high value units that can cause opponents great damage from a relatively longer distance.
Gun vehicles have less damage, shorter attack range but can withstand more damage during competitions;
Air drones possess high maneuverability, allowing them to navigate quickly and efficiently. They can be easily destoried under attack and can cause very little damage. Nevertheless, air drones are essential in LSOP because they are designed with the capability to repair damaged units nearby. An agent can be neutralized after it suffers from enough damage. Because of the variance in abilities, agents of a particular type have to develop policies that are not transferrable to other different agent types. 

LSOP is a two-team game that supports multiple configurations. LSOP-66x2 is a middle-scale configuration that is composed by 6 air drones, 12 missile vehicles and 48 gun vehicles at each team. LSOP-102x2 is a large-scale configuration with 6 air drones, 24 missile vehicles and 72 gun vehicles. Notably, the disparity in quantity among different types leads to varying levels of learning difficulty for different types of agents.

\textbf{Reward.} Each episode ends when time limit is reached or one team is eliminated entirely. The winner team, which has more remaining agents, will receive a positive reward $r$,while the defeated team will receive a negative reward $-r$.
The reward is $r = c \cdot \delta_N$, where $c = 0.1$ is a constant and $\delta_N$ is the difference of remaining number of agents between the winning team and the losing team.

The reward function is designed to be sparse, meaning that agents will receive no any rewards until an episode ends. The reward formulation promotes the emergence of sophisticated behaviors that are not influenced by human bias. Furthermore,
it is important to note that reward signals are given to teams instead of specific agents. In this way, we avoid introduce any assumptions considering the credit assignment, which can potentially introduce bias and cause undesired agent behaviors.

As LSOP has two teams in rivery, the algorithm controlling each team can be configurated differently. We has a expert-knowledge-based team controller as the baseline opponent. But LSOP also allows other SOTA algorithm to participant training or testing and play the role of opponent.
The expert baseline controller has the following hard-coded policy: agents are split into attack groups according to team composition. Considering air drone are the only agents with team-support capability, each air drone leads a attack group and acts as the group leader. The rest of the agents are distributed evenly to different attack groups. Next, each attack group evaluates opponent targets, determine which target has most threat, and then launch attack. The decisions are made in real time. Importantly, this baseline controller has full state observation and is allowed to make decisions in a centralized way. In other words, the baseline team controller is not restricted to CTDE paradigm, simplified the programming of agent behaviors and make it a potent opponent for MARL-based learners.

\begin{table}[t]
  \centering
  \caption{Task Environment Simulation Parameters.}
  \label{tab:env}
  \begin{tabular}{ccc}
  \toprule
  \textbf{Order} & \textbf{Task Parameters} & \textbf{Selection} \\
  \midrule
  1 & Simulation Step Time $dt_{sim}$  &  1/2560 \\ 
  2 & Action Step Time $dt_{sim}$  &  1/2 \\ 
  3 & Simulation Acceration  &  x64 \\ 
  4 & Parallel Simulation (Vector Env)  &  16 \textasciitilde \ 64 \\ 
  \bottomrule
  \end{tabular}
\end{table}

\subsection{Experimental Setup.}
Our experiments are conducted on a single Linux machine equipped with NVIDIA GPUs, specifically the RTX 8000 model. Only one GPU is utilized per experiment. In order to improve the training efficiency, we adopt a specific version of PPO with dual-clip \cite{ye2020mastering}. And Adam optimizer \cite{kingma2014adam} is used for training the parameters of actor and critic neural networks. The hyperparameters are listed in Table.~\ref{tab:hyper}.

\begin{table}[t]
  \centering
  \caption{Default hyperparameters selection in PHLRL.}
  \label{tab:hyper}
  \begin{tabular}{ccc}
  \toprule
  \textbf{Order} & \textbf{Hyperparameters} & \textbf{Selection} \\
  \midrule
  1 & League capacity   &  $K_m=10$\\ 
  2 & Batch size (Num of episodes) & $N_{batch}=32$\\ 
  3 & PPO iterations between league update & $N_u=100$\\ 
  4 & PPO steps per iteration & 24 \\ 
  5 & Num of Episodes during a single test  & 320 \\
  6 & Entropy loss coefficient  & 0.0005 \\ 
  7 & Critic Learning rate & 0.004 \\ 
  8 & Actor Learning rate  & 0.0004\\ 
  9 & Discount factor  & 0.99 \\
  10 & Reward Propogation Discount  & 0.98 \\
  \bottomrule
  \end{tabular}
\end{table}

\begin{figure*}[!t]
  \centering
  \begin{subfigure}[t]{0.18\linewidth}
    \centering
    \includegraphics[width=\linewidth,height=\linewidth]{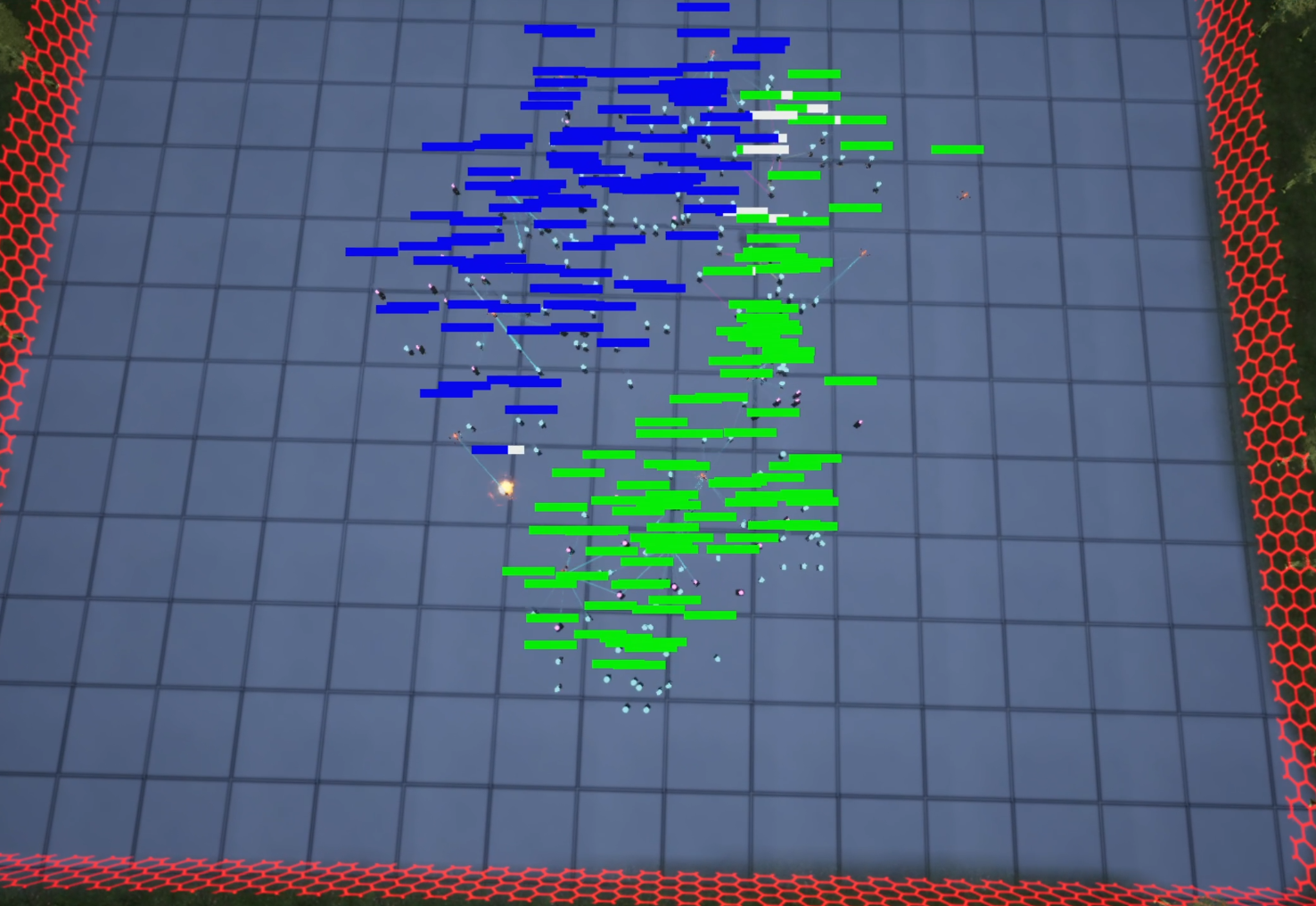}
    \caption{}
  \end{subfigure}
  \begin{subfigure}[t]{0.18\linewidth}
    \centering
    \includegraphics[width=\linewidth,height=\linewidth]{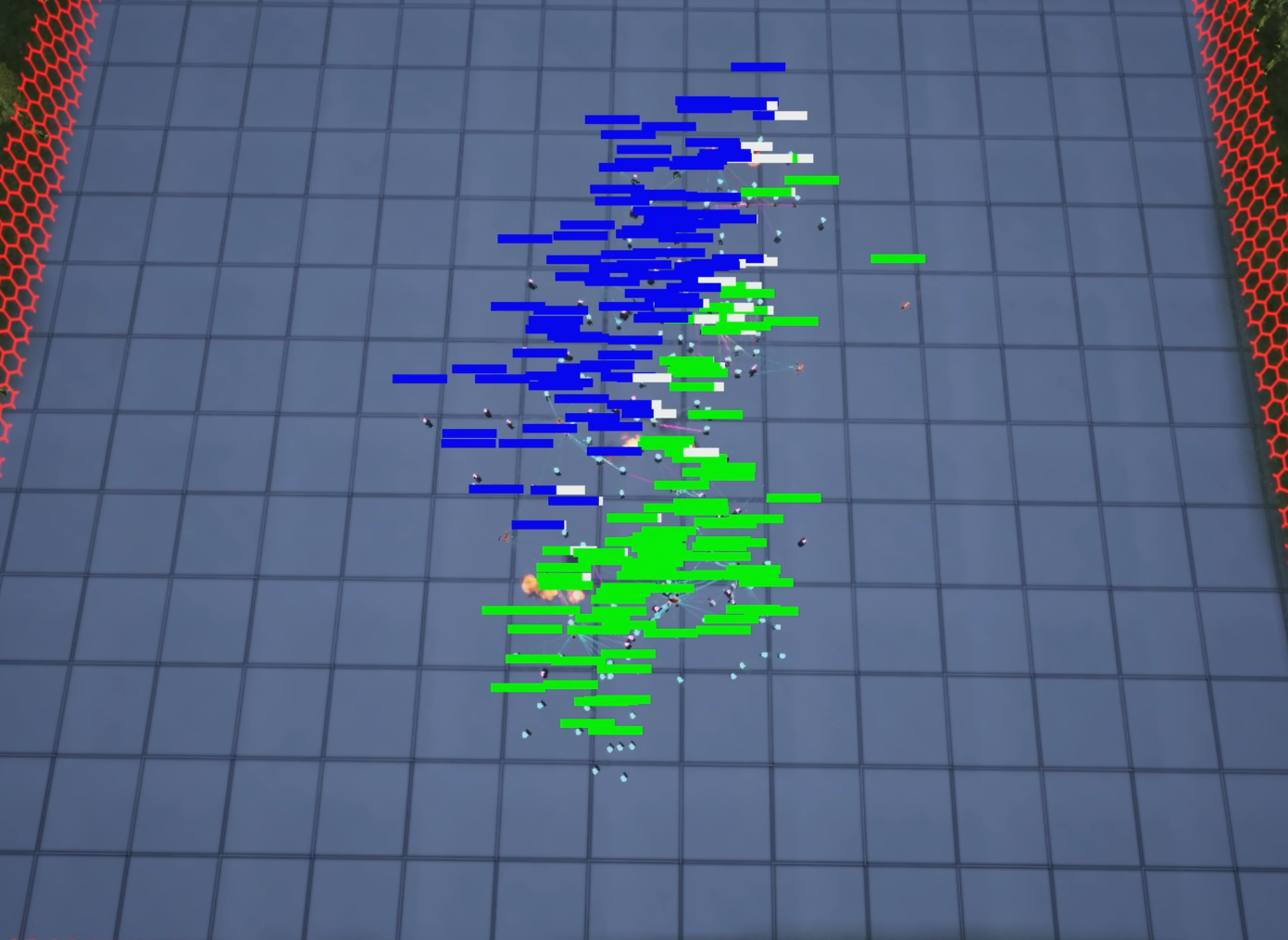}
    \caption{}
  \end{subfigure}
  \begin{subfigure}[t]{0.18\linewidth}
    \centering
    \includegraphics[width=\linewidth,height=\linewidth]{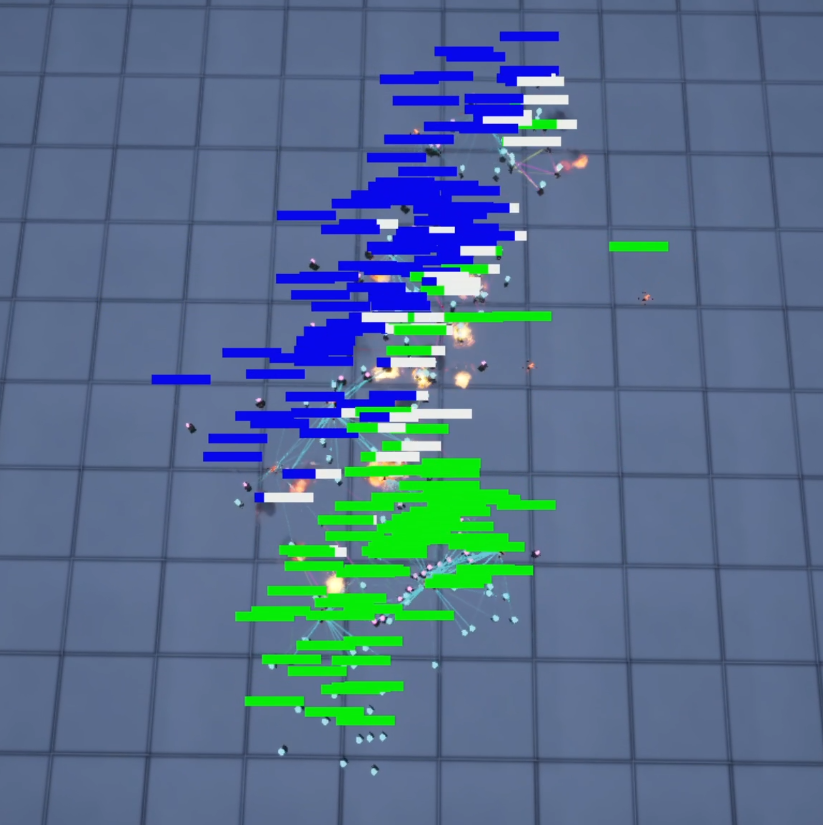}
    \caption{}
  \end{subfigure}
  \begin{subfigure}[t]{0.18\linewidth}
    \centering
    \includegraphics[width=\linewidth,height=\linewidth]{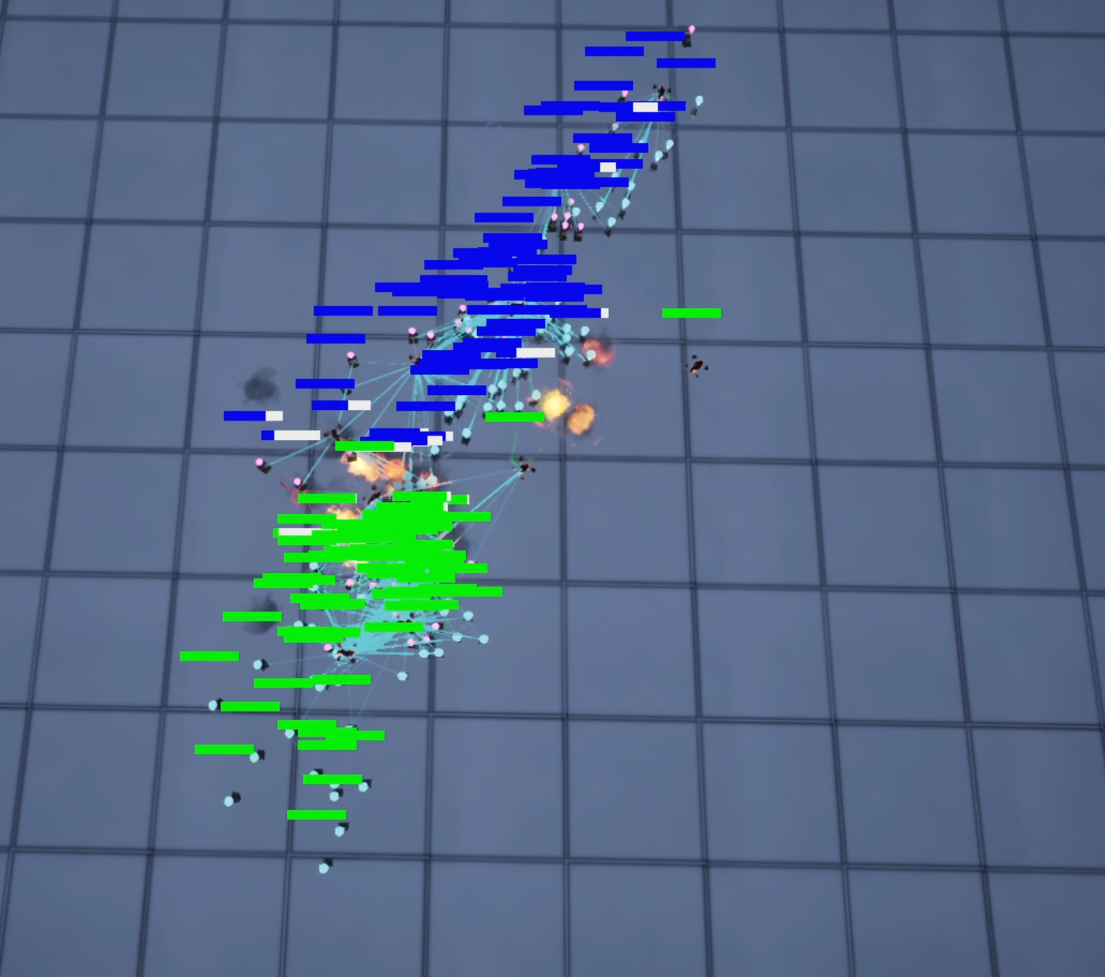}
    \caption{}
  \end{subfigure}\\
  \begin{subfigure}[t]{0.18\linewidth}
    \centering
    \includegraphics[width=\linewidth,height=\linewidth]{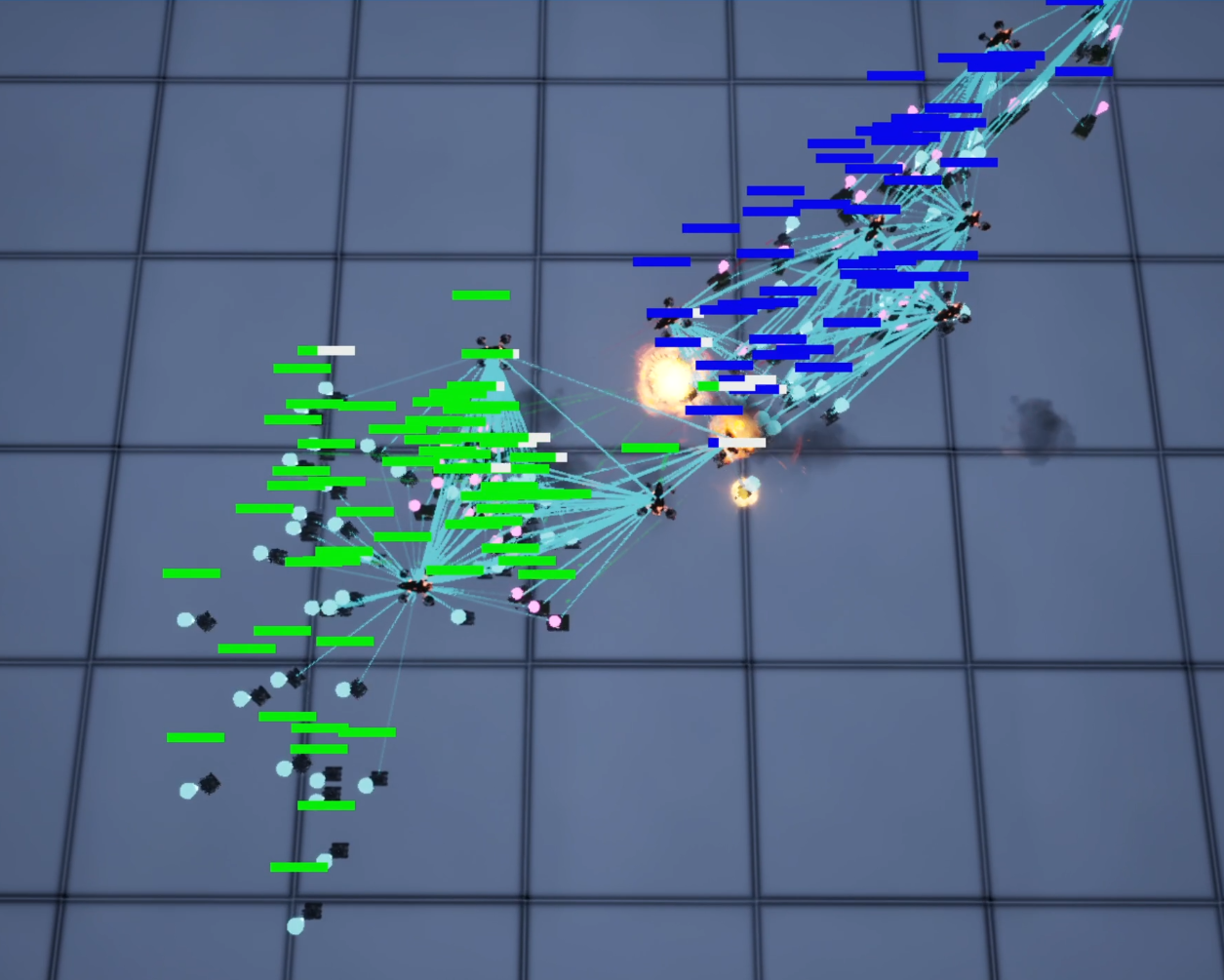}
    \caption{}
  \end{subfigure}
  \begin{subfigure}[t]{0.18\linewidth}
    \centering
    \includegraphics[width=\linewidth,height=\linewidth]{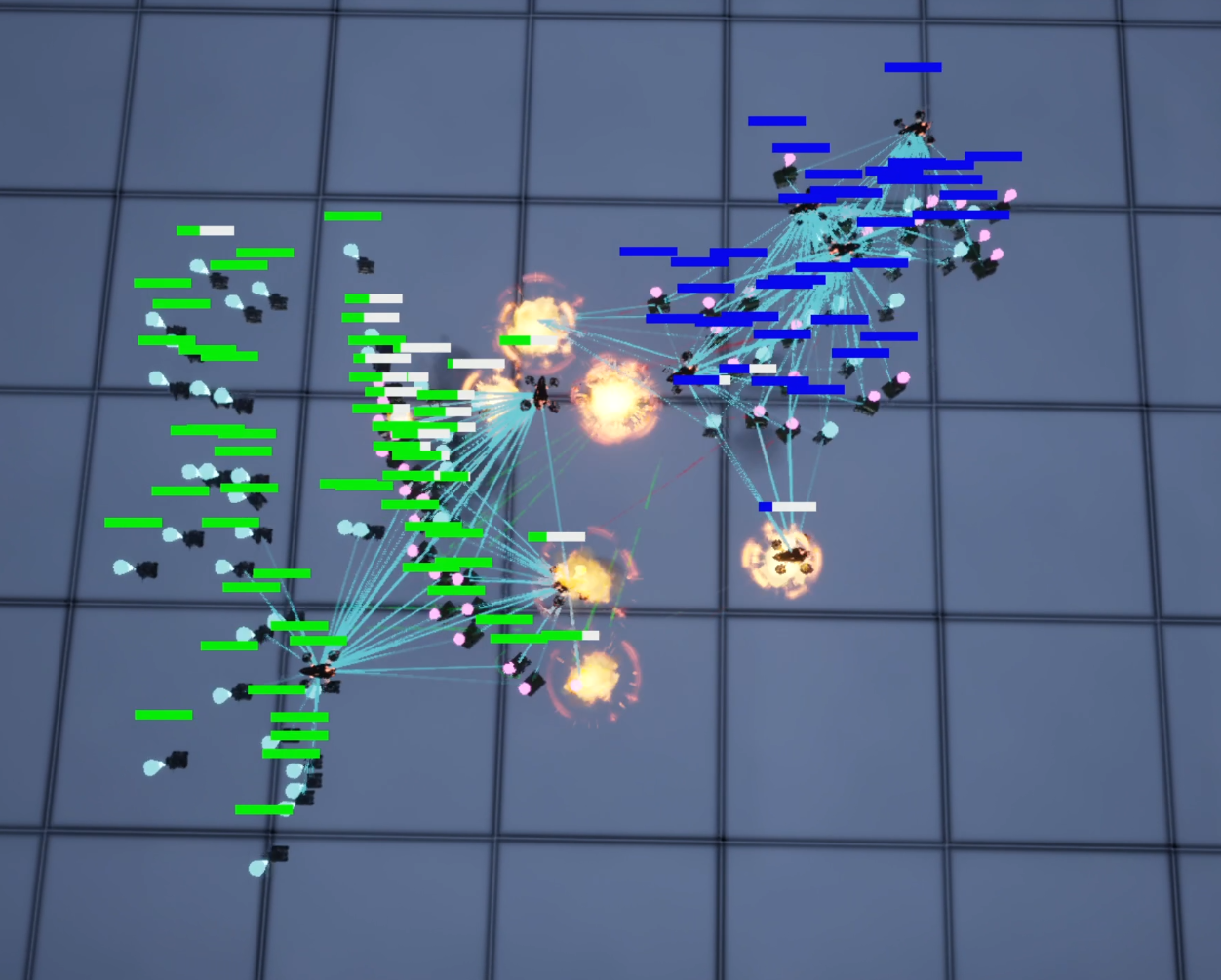}
    \caption{}
  \end{subfigure}
  \begin{subfigure}[t]{0.18\linewidth}
    \centering
    \includegraphics[width=\linewidth,height=\linewidth]{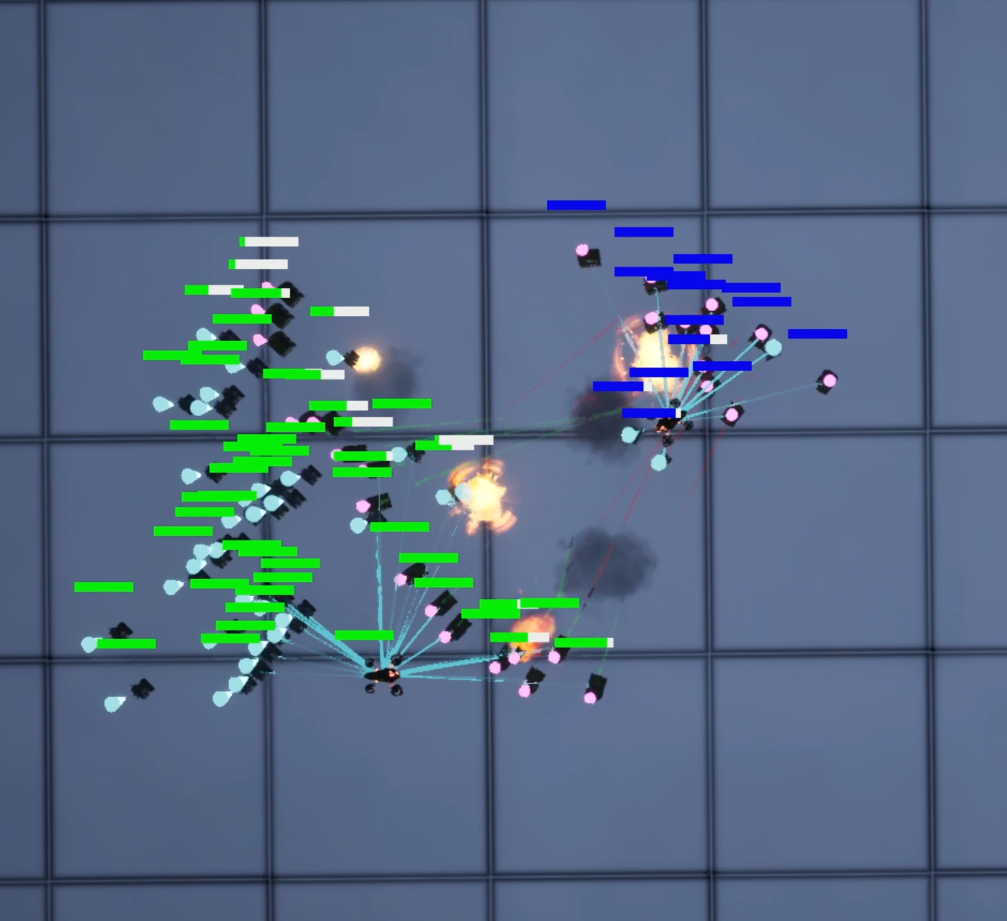}
    \caption{}
  \end{subfigure}
  \begin{subfigure}[t]{0.18\linewidth}
    \centering
    \includegraphics[width=\linewidth,height=\linewidth]{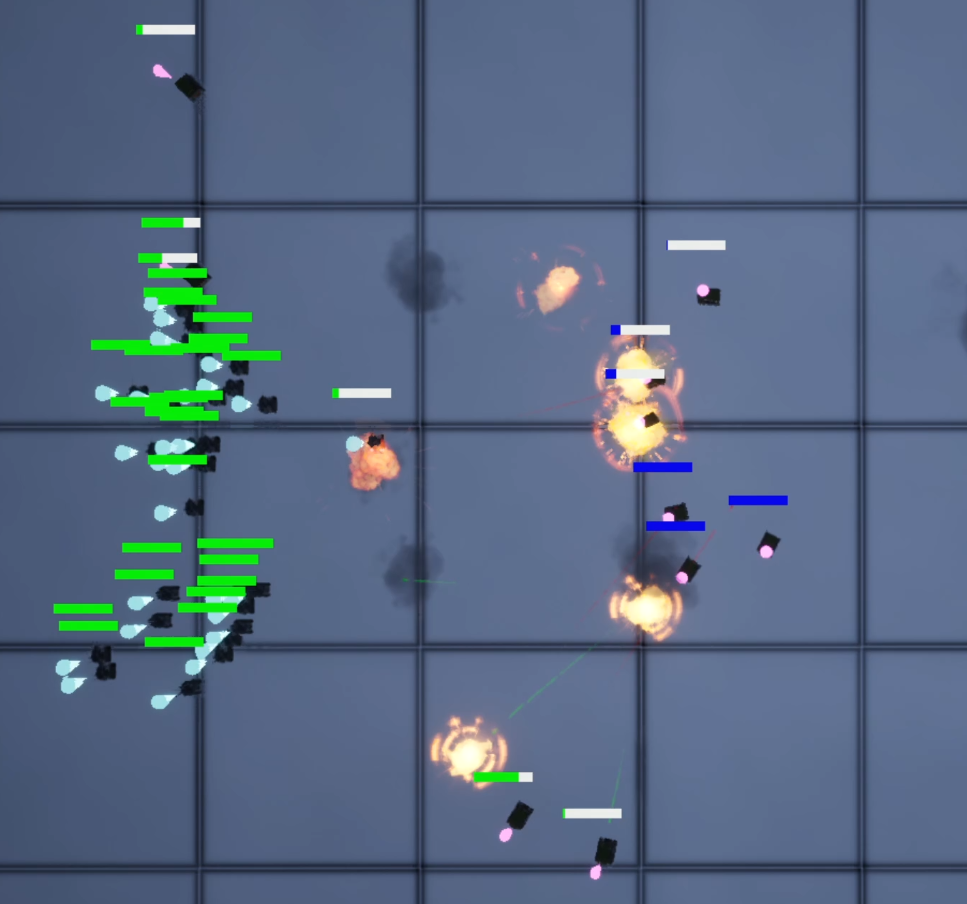}
    \caption{}
  \end{subfigure}
  \caption{
    Detour heterogeneous cooperative policy learned by PHLRL in LSOP. PHLRL agents are in green color, while their opponents are in blue color.
  }
\label{fig:LSOP1}
\end{figure*}

\begin{figure*}[!t]
  \centering
  \begin{subfigure}[t]{0.22\linewidth}
    \centering
    \includegraphics[width=\linewidth,height=\linewidth]{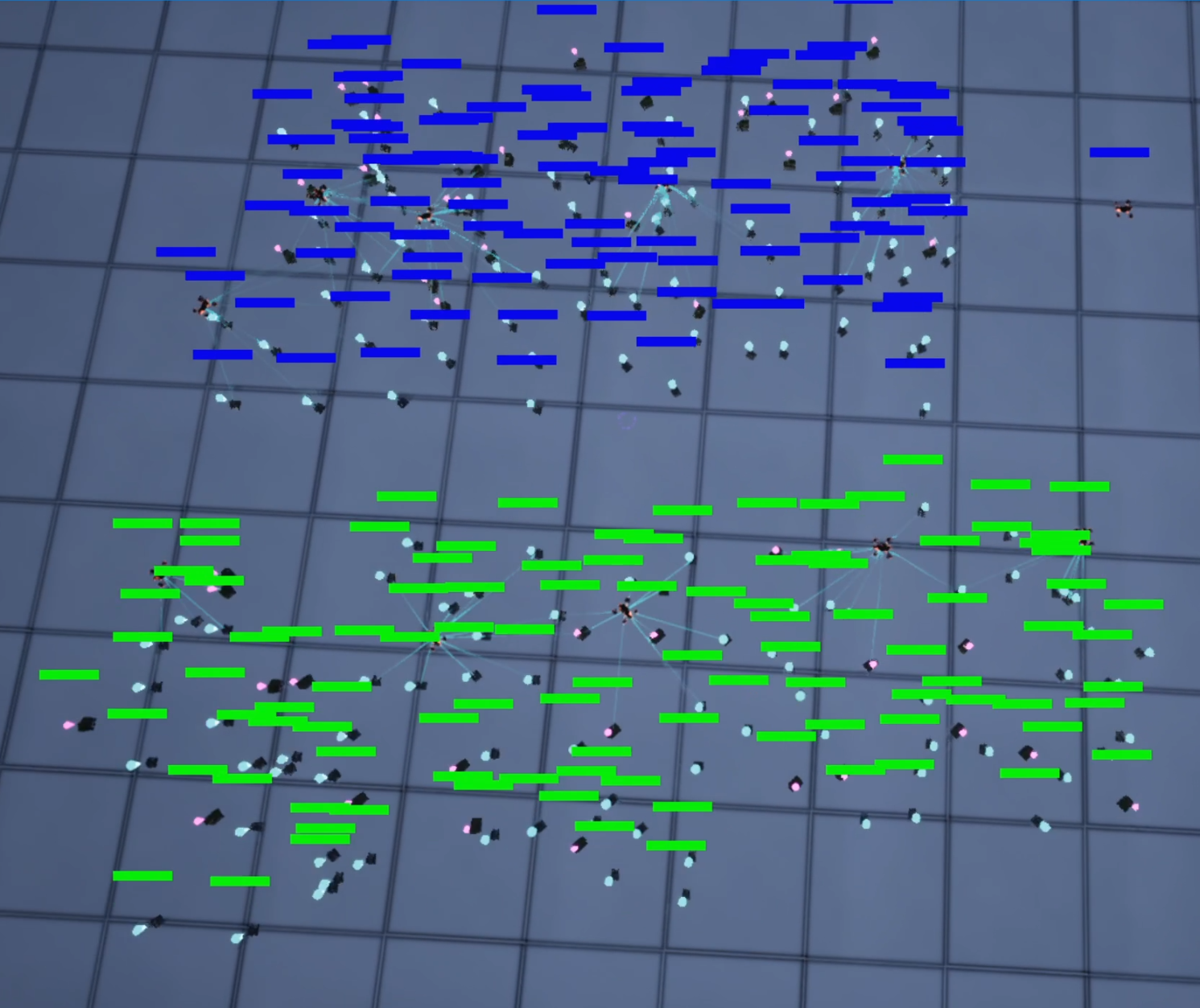}
    \caption{}
  \end{subfigure}
  \begin{subfigure}[t]{0.22\linewidth}
    \centering
    \includegraphics[width=\linewidth,height=\linewidth]{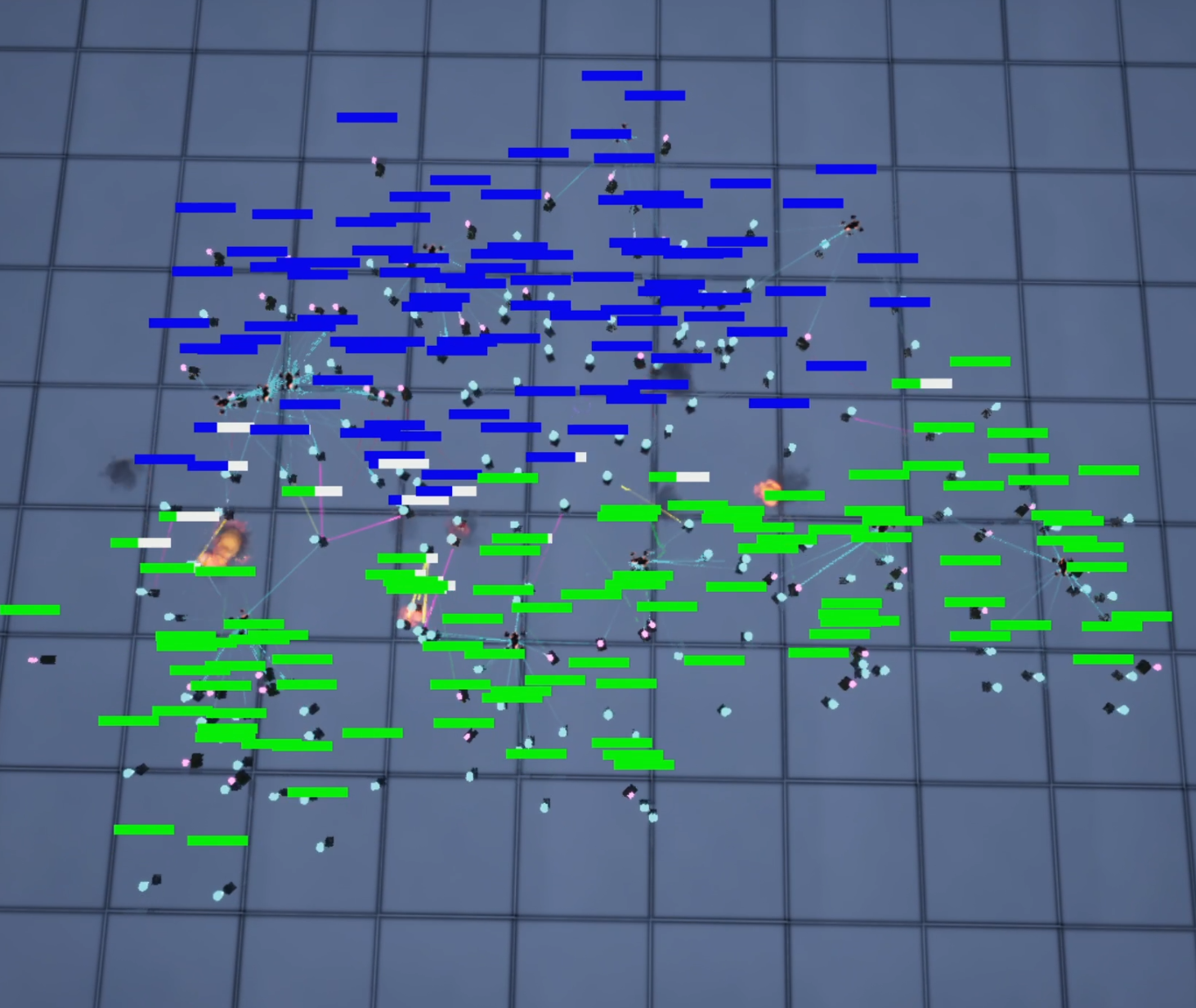}
    \caption{}
  \end{subfigure}
  \begin{subfigure}[t]{0.22\linewidth}
    \centering
    \includegraphics[width=\linewidth,height=\linewidth]{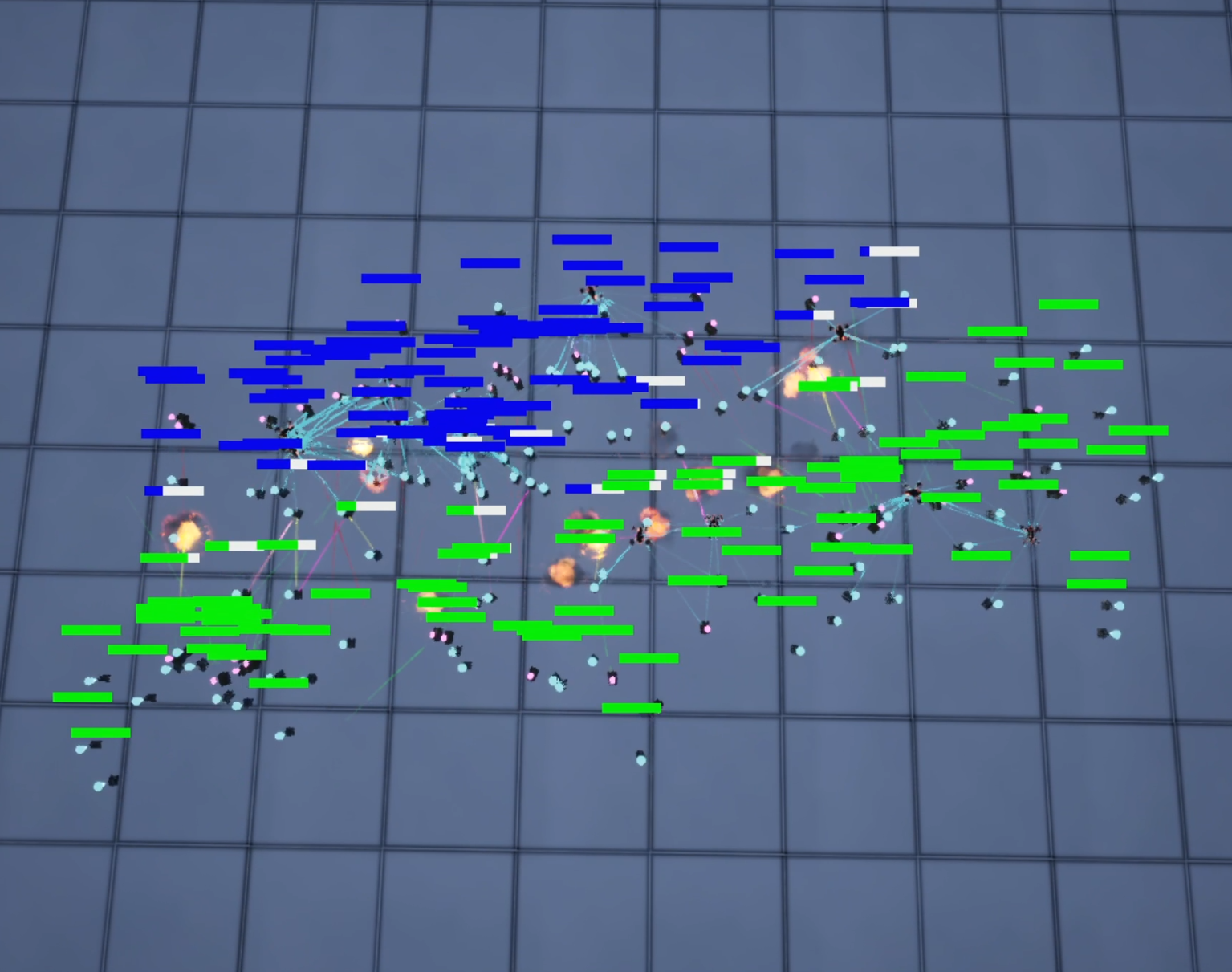}
    \caption{}
  \end{subfigure}\\
  \begin{subfigure}[t]{0.22\linewidth}
    \centering
    \includegraphics[width=\linewidth,height=\linewidth]{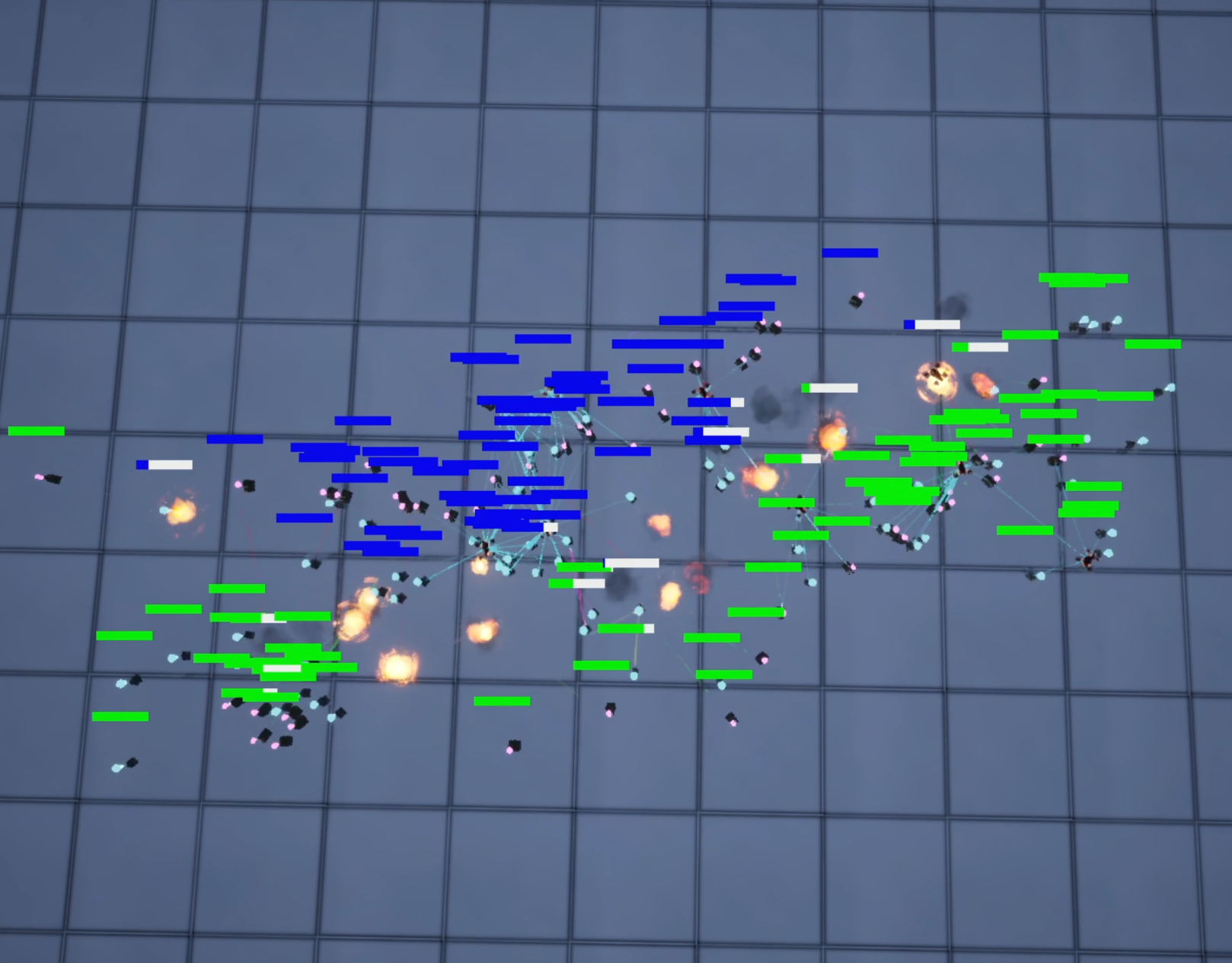}
    \caption{}
  \end{subfigure}
  \begin{subfigure}[t]{0.22\linewidth}
    \centering
    \includegraphics[width=\linewidth,height=\linewidth]{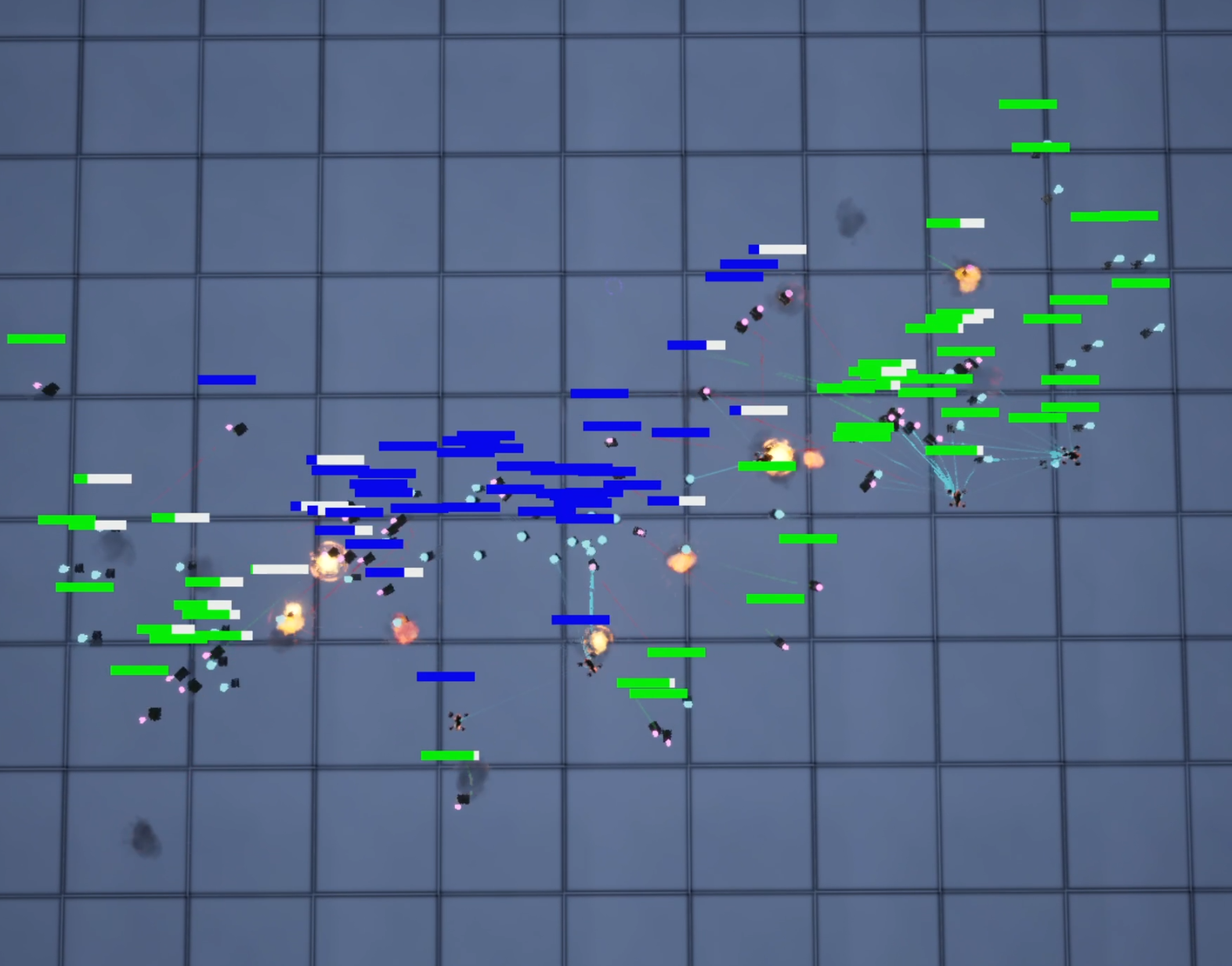}
    \caption{}
  \end{subfigure}
  \begin{subfigure}[t]{0.22\linewidth}
    \centering
    \includegraphics[width=\linewidth,height=\linewidth]{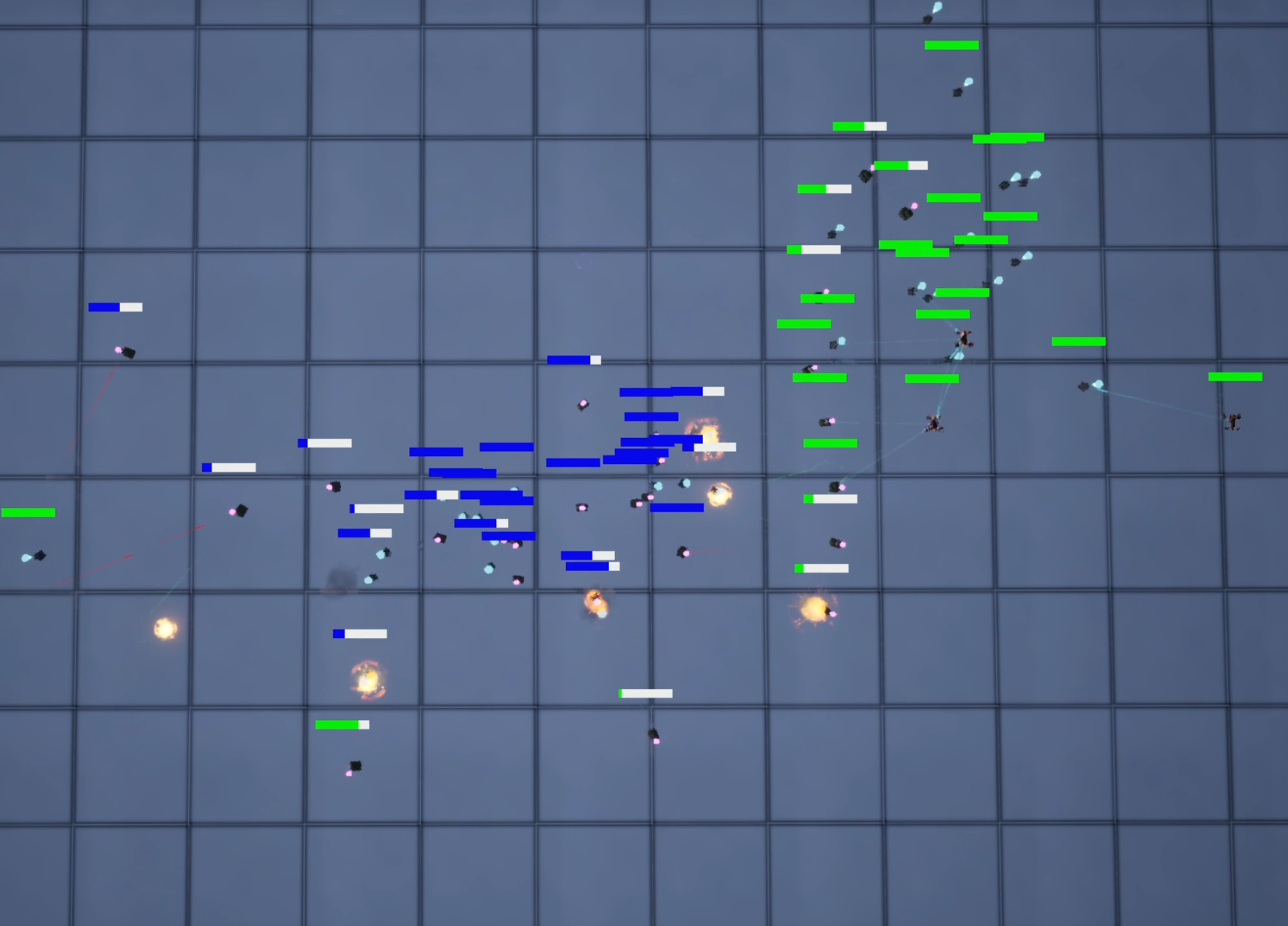}
    \caption{}
  \end{subfigure}
  \caption{
    Direct confrontation and flanking policy learned by PHLRL in LSOP. PHLRL agents are in green color, while their opponents are in blue color.
  }
\label{fig:LSOP2}
\end{figure*}


\begin{figure}[!t]
\begin{subfigure}[t]{0.8\linewidth}
    \centering
    \includegraphics[width=\linewidth]{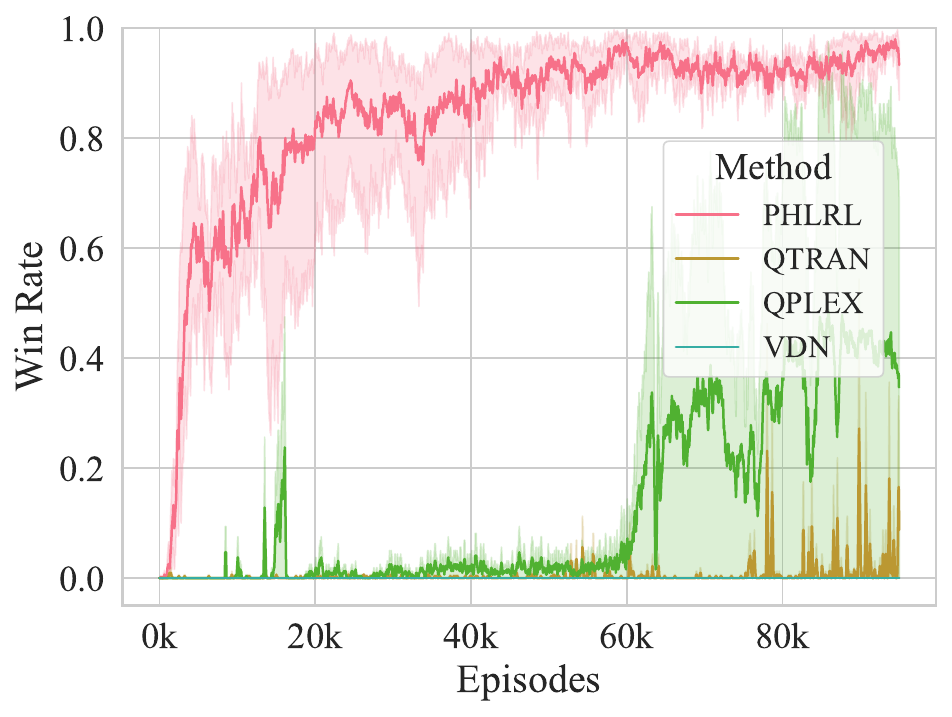}
\end{subfigure}
\begin{subfigure}[t]{0.8\linewidth}
  \centering
  \includegraphics[width=\linewidth]{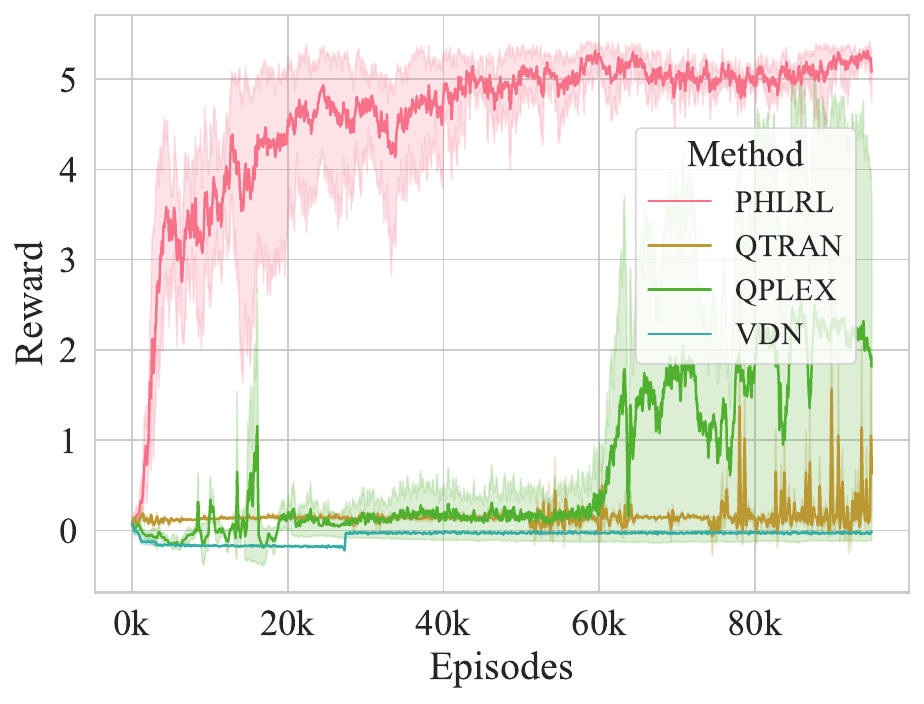}
\end{subfigure}
\caption{
  The performance of PHLRL and other MARL methods including QTRAN, QPLEX, CW-QMIX and VDN.
}
\label{fig:performance}
\end{figure}

\begin{figure}[!t]
  \centering
  \includegraphics[width=0.8\linewidth]{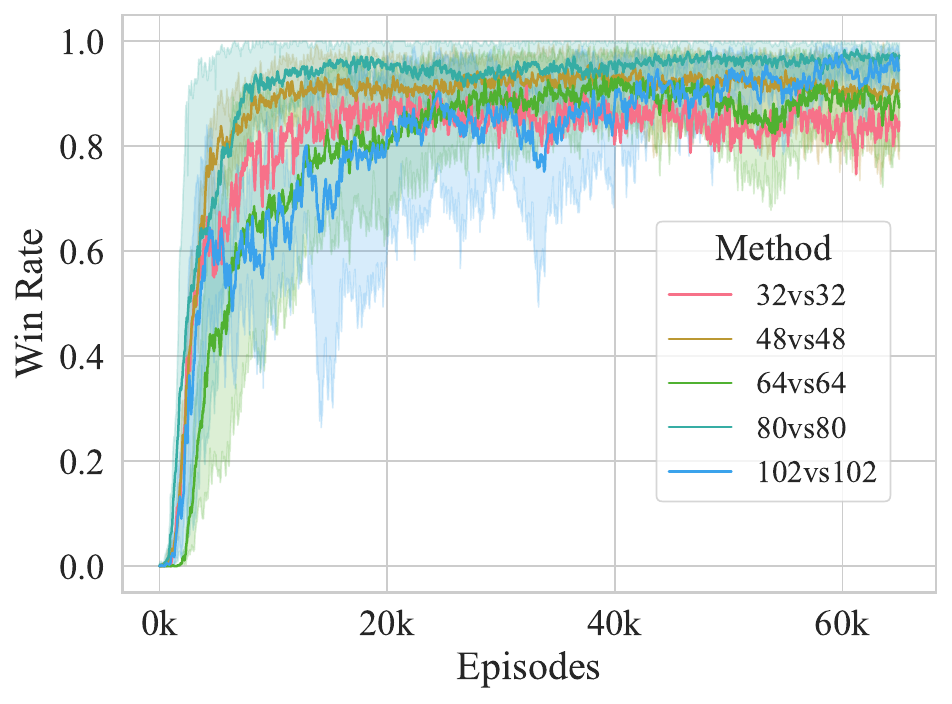}
  \caption{Testing the scalability of PHLRL.}
  \label{fig:scalability}
  \end{figure}

\subsection{Overall Performance of PHLRL}

The performance of PHLRL is demonstrated in Fig.~\ref{fig:performance}. In this experiment, the PHLRL is trained to compete with the baseline expert policy. Secondly, the PHLRL method is compared with several SOTA MARL algorithms including QPLEX \cite{wang2020qplex}, VDN \cite{su2021value} and QTRAN \cite{son2019qtran}.

\subsubsection{The performance of PHLRL}
According to Fig.\ref{fig:performance}, the PHLRL method has a significant advantage over the baseline expert policy. The win rate of PHLRL is above 90\% in most experiments within 60k epsiodes. On the other hand, the averaged win rate of QPLEX is less than 50\%, the average win rate of QTRAN is round 30\% and VDN fails to learn effective cooperative policy (ends up with 0\% win rate) in this test environment.
This result demonstrates the effectiveness of PHLRL among existing methods in heterogeneous multiagent tasks.

\subsubsection{PHLRL Policies}
The cooperative policies of PHLRL agents learned in difference experiments have shown diversity.
Fig.~\ref{fig:LSOP} and Fig.~\ref{fig:LSOP2} illustrate two distinct strategies learned in two PHLRL experiments with same setting except for the random seed. 

Fig.~\ref{fig:LSOP} gives a demonstration of a detour cooperative strategy. Agents at the initial training stage are unable to defeat the opposing team by direct confrontation, therefore the team adopts a fight-and-maneuver strategy. Firstly, most agent move south to change the formation from a long line to a densed cluster (Fig.~\ref{fig:LSOP}-a\textasciitilde d). However, the re-formation process is risky as the whole team becomes vunerable to opponent attacks,
the team learns to resolve this problem by sending a small group of agents to distract the opponents and provide covering fire for team reformation.
These agents carrying out the distraction task suffer great lost (Fig.~\ref{fig:LSOP}-c), however, during this process the main force of the team successfully gathers and shifts to the weak point of the enemy front (Fig.~\ref{fig:LSOP}-d). 
When the local advantage is established at the enemy front, the team seize the chance to attack the opponent agents and create a good exchange ratio of casualties that is in favor of PHLRL agents (Fig.~\ref{fig:LSOP}-e).
By the time the opponent re-adjust their formation, PHLRL agents have already achieved leading position (Fig.~\ref{fig:LSOP}-f) and can consequently win the game in the end (Fig.~\ref{fig:LSOP}-e\textasciitilde h).

Fig.~\ref{fig:LSOP2} illustrates another strategy featured by direct confrontation and encirclement.
Instead of avoid initial conflicts, agents start by commiting direct attack in this instance of experiment.
Nevertheless,
the densely packed formation heightens the risk of enemy missile attacks, as the blast radius encompasses all targets within its reach. Conversely, a sparse formation may lead to a scenario where agents positioned too far from the front line are incapacitated from launching any attack on the enemy.
PHLRL agents resolve this problem by using a flanking movement to encircle their opponents (Fig.~\ref{fig:LSOP2}-c\textasciitilde e).
Moreover, the flanking cooperative policy significantly limit the space of maneuvering for the opponent team, which makes it difficult for them to re-adjust their formation and launches counter-attack.
This strategy can effectively improve the exchange ratio of casualties and achieve higher win rate.




\subsection{Scalability.}

In this section the scalability of PHLRL is tested by changing the number of agents in both team. Other scenarios with distinct scales are tested, including 32vs32, 48vs48, 64vs64 and 80vs80. The results are shown in Fig.~\ref{fig:scalability}. The result suggests that in all settings our PHLRL method can achieve success rate above 80\%. Specially, due to the application of prioritized policy gradient, our method can achieve even better performance in 80vs80 and 102vs102 scenarios. 
In both scenarios the averaged win rate of PHLRL can reach 90\% within 60k epsiodes.

These experiments demonstrate that PHLRL

\section{Conclusions}

This paper proposes a Prioritized Heterogeneous League Reinforcement Learning (PHLRL) method, which aims to address cooperative heterogeneous multiagent challenges in large-scale teams.
PHLRL method is featured by a heterogeneous league that promotes agents' cooperation capability between different agent types and a prioritized policy gradient training framework that resolves the heterogeneous sample inequality problem.
The effectiveness of PHLRL is compared with multiple SOTA methods and PHLRL has demonstrated better performance in large-scale heterogeneous games.
Finally, the experiments show that PHLRL has the scalability to be extended and solve heterogeneous problems with different scales.

\bibliographystyle{IEEEtranN}
{\footnotesize
\bibliography{cite.bib}}



\newpage

\vfill

\end{document}